\documentclass{article}

\usepackage{multirow}

\usepackage[final]{neurips_2023}

\usepackage[utf8]{inputenc} %
\usepackage[T1]{fontenc}    %
\usepackage{url}            %
\usepackage{booktabs}       %
\usepackage{amsfonts}       %
\usepackage{nicefrac}       %
\usepackage{microtype}      %
\usepackage{xcolor}         %
\usepackage{amsmath}
\usepackage{amssymb}
\usepackage{mathtools}
\usepackage{slashed}
\usepackage{braket}
\usepackage{enumitem}
\usepackage[colorlinks,citecolor=darkgreen,linkcolor=firebrick,urlcolor=firebrick]{hyperref}
\usepackage{float}

\usepackage{graphicx}
\usepackage{tikz}
\usepackage{tikz-cd}
\usepackage{times}
\usepackage{courier}
\usepackage{bm}
\usepackage{subfig}
\usepackage{physics}
\usepackage{xcolor}
\usepackage{natbib}
\usepackage{mdframed}
\usepackage{nicefrac}
\usepackage{booktabs}
\usepackage{lipsum}
\usepackage{titlesec}
\usepackage{wrapfig,lipsum,booktabs}
\usepackage{authblk}
\usepackage{blindtext}
\usepackage{algorithm}
\usepackage{algorithmicx}
\usepackage{algpseudocode}
\usepackage{titletoc}
\usepackage{todonotes}
\usepackage{authblk}
\usepackage{titletoc}

\usepackage[capitalize,noabbrev]{cleveref}

\usepackage{CJK}

\renewenvironment{figure}[1][]{
  \begin{originalfigure}[#1]
    \begin{mdframed}[linecolor=black!0,backgroundcolor=black!1]
}{
    \end{mdframed}
  \end{originalfigure}
}

\usepackage{array}
\usepackage{makecell}

\newcommand{\bea}{\begin{eqnarray}}
\newcommand{\eea}{\end{eqnarray}}

\usepackage{slashed}

\def\({\left(}
\def\){\right)}
\def\[{\left[}
\def\]{\right]}

\usepackage{dashbox}
\usepackage{xcolor}
\usepackage{colortbl}
\usepackage{arydshln}
\definecolor{lightyellow}{rgb}{1.0, 0.95, 0.7}
\definecolor{blue}{rgb}{0.0, 0.4, 1.0}
\definecolor{Blue}{rgb}{0,0,1}
\definecolor{darkgreen}{rgb}{0,0.40,0}
\definecolor{firebrick}{rgb}{0.698,0.133,0.133}

\definecolor{colorA}{rgb}{1,0,0}
\definecolor{colorB}{rgb}{0,0.3,1}
\definecolor{colorC}{rgb}{0.9,0.8,0.2}
\definecolor{colorD}{rgb}{0,0.65,0}
\definecolor{lesslightgray}{rgb}{0.5,0.5,0.5}
\definecolor{light-gray}{gray}{0.95}

\let\cite\citep 

\usepackage{amsthm}
\makeatletter
\def\th@remark{%
  \thm@headfont{\bfseries}%
  \normalfont %
  \thm@preskip\topsep \divide\thm@preskip\tw@
  \thm@postskip\thm@preskip
}
\makeatother
\usepackage[many]{tcolorbox}
\theoremstyle{definition}

\tcolorboxenvironment{theorem}{
  breakable,
  colback=black!10,
  colframe=white,%
  width=\dimexpr\linewidth+10pt\relax,%
  enlarge left by=-5pt,%
  enlarge right by=-5pt,%
  boxsep=5pt,%
  boxrule=0pt,
  left=0pt,right=0pt,top=0pt,bottom=0pt,
  sharp corners,
  before skip=\topsep,
  after skip=\topsep
}

\tcolorboxenvironment{lemma}{
  breakable,
  colback=black!10,
  colframe=white,%
  width=\dimexpr\linewidth+10pt\relax,%
  enlarge left by=-5pt,%
  enlarge right by=-5pt,%
  boxsep=5pt,%
  boxrule=0pt,
  left=0pt,right=0pt,top=0pt,bottom=0pt,
  sharp corners,
  before skip=\topsep,
  after skip=\topsep
}

\tcolorboxenvironment{corollary}{
  breakable,
  colback=black!10,
  colframe=white,%
  width=\dimexpr\linewidth+10pt\relax,%
  enlarge left by=-5pt,%
  enlarge right by=-5pt,%
  boxsep=5pt,%
  boxrule=0pt,
  left=0pt,right=0pt,top=0pt,bottom=0pt,
  sharp corners,
  before skip=\topsep,
  after skip=\topsep
}

\theoremstyle{definition}

\tcolorboxenvironment{definition}{
  breakable,
  colback=black!10,
  colframe=white,%
  width=\dimexpr\linewidth+10pt\relax,%
  enlarge left by=-5pt,%
  enlarge right by=-5pt,%
  boxsep=5pt,%
  boxrule=0pt,
  left=0pt,right=0pt,top=0pt,bottom=0pt,
  sharp corners,
  before skip=\topsep,
  after skip=\topsep
}
\theoremstyle{remark}

\tcolorboxenvironment{assumption}{
  breakable,
  colback=black!10,
  colframe=white,%
  width=\dimexpr\linewidth+10pt\relax,%
  enlarge left by=-5pt,%
  enlarge right by=-5pt,%
  boxsep=5pt,%
  boxrule=0pt,
  left=0pt,right=0pt,top=0pt,bottom=0pt,
  sharp corners,
  before skip=\topsep,
  after skip=\topsep
}

\tcolorboxenvironment{problem}{
  breakable,
  colback=black!10,
  colframe=white,%
  width=\dimexpr\linewidth+10pt\relax,%
  enlarge left by=-5pt,%
  enlarge right by=-5pt,%
  boxsep=5pt,%
  boxrule=0pt,
  left=0pt,right=0pt,top=0pt,bottom=0pt,
  sharp corners,
  before skip=\topsep,
  after skip=\topsep
}

\crefname{theorem}{Theorem}{Theorems}
\crefname{proposition}{Proposition}{Propositions}
\crefname{lemma}{Lemma}{Lemmas}
\crefname{corollary}{Corollary}{Corollaries}
\crefname{definition}{Definition}{Definitions}
\crefname{assumption}{Assumption}{Assumptions}
\crefname{remark}{Remark}{Remarks}
\crefname{problem}{Problem}{Problems}
\crefname{property}{Property}{property}

\numberwithin{equation}{section}
\numberwithin{theorem}{section}
\numberwithin{proposition}{section}
\numberwithin{definition}{section}
\numberwithin{lemma}{section}
\numberwithin{assumption}{section}
\numberwithin{remark}{section}

\usepackage{lipsum}

\makeatletter
\let\save@mathaccent\mathaccent
\newcommand*\if@single[3]{%
    \setbox0\hbox{${\mathaccent"0362{#1}}^H$}%
    \setbox2\hbox{${\mathaccent"0362{\kern0pt#1}}^H$}%
    \ifdim\ht0=\ht2 #3\else #2\fi
}
\newcommand*\rel@kern[1]{\kern#1\dimexpr\macc@kerna}
\newcommand*\widebar[1]{\@ifnextchar^{{\wide@bar{#1}{0}}}{\wide@bar{#1}{1}}}
\newcommand*\wide@bar[2]{\if@single{#1}{\wide@bar@{#1}{#2}{1}}{\wide@bar@{#1}{#2}{2}}}
\newcommand*\wide@bar@[3]{%
    \begingroup
    \def\mathaccent##1##2{%
        \let\mathaccent\save@mathaccent
        \if#32 \let\macc@nucleus\first@char \fi
        \setbox\z@\hbox{$\macc@style{\macc@nucleus}_{}$}%
        \setbox\tw@\hbox{$\macc@style{\macc@nucleus}{}_{}$}%
        \dimen@\wd\tw@
        \advance\dimen@-\wd\z@
        \divide\dimen@ 3
        \@tempdima\wd\tw@
        \advance\@tempdima-\scriptspace
        \divide\@tempdima 10
        \advance\dimen@-\@tempdima
        \ifdim\dimen@>\z@ \dimen@0pt\fi
        \rel@kern{0.6}\kern-\dimen@
        \if#31
        \overline{\rel@kern{-0.6}\kern\dimen@\macc@nucleus\rel@kern{0.4}\kern\dimen@}%
        \advance\dimen@0.4\dimexpr\macc@kerna
        \let\final@kern#2%
        \ifdim\dimen@<\z@ \let\final@kern1\fi
        \if\final@kern1 \kern-\dimen@\fi
        \else
        \overline{\rel@kern{-0.6}\kern\dimen@#1}%
        \fi
    }%
    \macc@depth\@ne
    \let\math@bgroup\@empty \let\math@egroup\macc@set@skewchar
    \mathsurround\z@ \frozen@everymath{\mathgroup\macc@group\relax}%
    \macc@set@skewchar\relax
    \let\mathaccentV\macc@nested@a
    \if#31
    \macc@nested@a\relax111{#1}%
    \else
    \def\gobble@till@marker##1\endmarker{}%
    \futurelet\first@char\gobble@till@marker#1\endmarker
    \ifcat\noexpand\first@char A\else
    \def\first@char{}%
    \fi
    \macc@nested@a\relax111{\first@char}%
    \fi
    \endgroup
    }
\makeatother

\makeatletter
\newcommand*{\redefinesymbolwitharg}[1]{%
  \expandafter\let\csname ltx#1\expandafter\endcsname\csname #1\endcsname
  \@namedef{#1}{\@ifnextchar{^}{\@nameuse{#1@}}{\@nameuse{#1@}^{}}}%
  \expandafter\def\csname #1@\endcsname^##1##2{%
     \csname ltx#1\endcsname\ifx!##1!\else^{##1}\fi\mathopen{}\mathclose\bgroup\left(##2\aftergroup\egroup\right)
     }%
}
\makeatother
\redefinesymbolwitharg{sin}
\redefinesymbolwitharg{cos}

\usepackage{titletoc}

\titlespacing*{\section}{0pt}{0pt}{0pt}
\titlespacing*{\subsection}{0pt}{0pt}{0pt}
\titlespacing*{\subsubsection}{0pt}{0pt}{0pt}

\setlist[itemize]{leftmargin=2em, before=\vspace{-0.5em}, after=\vspace{-0.5em}, itemsep=0.1em}
\title{Beyond PID Controllers:
PPO with Neuralized PID Policy  for  Proton Beam Intensity Control in Mu2e 
}

\author{
    C. Xu\thanks{Equal contribution}\;,                                   %
    J.YC. Hu$^*$,                                 %
    J. Jiang,                                 %
    S. Memik,                               %
    R. Shi,                                 %
    A.M. Shuping,                           %
    M. Thieme, H. Liu,                                 %
    \\ \vspace{-1em}
    Northwestern University\thanks{Performed at Northwestern with support from the Departments of Computer Science and Electrical and Computer Engineering}, Evanston, IL USA \\
    \vspace{0.5em}
    M.R. Austin,                            %
    J.M. Arnold,                            %
    J.R. Berlioz,                            %
    P. Hanlet,                              %
    K.J. Hazelwood,                          %
    M.A. Ibrahim,                           %
    J.St. John,                                   %
    J. Mitrevski,                           %
    V.P. Nagaslaev,                         %
    D.J. Nicklaus,                          %
    G. Pradhan,                             %
    A.L. Saewert,                           %
    B.A. Schupbach,                         %
    K. Seiya,                               %
    R.M. Thurman-Keup,                      %
    N.V. Tran,                              %
    A. Narayanan, \\     %
    Fermi National Accelerator Laboratory\thanks{Operated by Fermi Research Alliance, LLC under Contract No.De-AC02-07CH11359 with the United States Department of Energy. Additional funding provided by Grant Award No. LAB 20-2261 \cite{doe_lab_foa_2261}}, Batavia, IL USA 
}

\begin{document}

\maketitle
\vspace{-2em}
\begin{abstract}
    \vspace{-0.1in}
We introduce a novel Proximal Policy Optimization (PPO) algorithm aimed at addressing the challenge of maintaining a uniform proton beam intensity delivery in the \textbf{Mu}on \textbf{to} \textbf{E}lectron Conversion Experiment (\textbf{Mu2e}) at Fermi National Accelerator Laboratory (Fermilab). 
Our primary objective is to regulate the spill process to ensure a consistent intensity profile, with the ultimate goal of creating an automated controller capable of providing real-time feedback and calibration of the \textbf{S}pill \textbf{R}egulation \textbf{S}ystem (\textbf{SRS}) parameters on a millisecond timescale.
We treat the Mu2e accelerator system as a Markov Decision Process suitable for Reinforcement Learning (RL), utilizing PPO to reduce bias and enhance training stability. 
A key innovation in our approach is the integration of neuralized \textbf{P}roportional-\textbf{I}ntegral-\textbf{D}erivative (\textbf{PID}) controller into the policy function, resulting in a significant improvement in the \textbf{S}pill \textbf{D}uty \textbf{F}actor (\textbf{SDF}) by 13.6\%, surpassing the performance of the current PID controller baseline by an additional 1.6\%. 
This paper presents the preliminary offline results based on a differentiable simulator of the Mu2e accelerator.
It paves the ground works for real-time implementations and applications, representing a crucial step towards automated proton beam intensity control for the Mu2e experiment.
\end{abstract}

\section{Introduction}
\label{sec:intro}
\vspace{-0.05in}
We propose a novel RL-enhanced spill regularization system that incorporates a neuralized PID policy function to tackle the beam regularization challenge in the Mu2e experiments~\cite{bartoszek2015mu2e} at Fermilab. 
Our objective is to create an automated controller that ensures consistent spill (proton beams) intensity during experiments meeting real-time control requirements \cite{narayanan2021optimizing}.
This objective falls under the broader scope of the Accelerator \textbf{R}eal-time \textbf{E}dge \textbf{A}I for \textbf{D}istributed \textbf{S}ystems (\textbf{READS}) project \cite{mitrevski-fmlfs23,seiya:reads-foa,hazelwood:ipac21-mopab288}.
To achieve this, we model the Mu2e accelerator system as a Markov Decision Process and employ the Proximal Policy Optimization (PPO) algorithm \cite{schulman2017proximal} to cast the spill regulations as sequential decision-making problems. 
Our main contribution is the integration of a neuralized PID policy function \cite{zribi2018new} for our RL framework, encompassing the inductive bias of the standard PID controller (i.e. the proportional, integral, and derivative information) to better capture states at different stages. 
Our experiments on the Mu2e simulator show that we observed an average improvement of 13.6\% in the \textbf{S}pill \textbf{D}uty \textbf{F}actor (\textbf{SDF}). 
Additionally, our method outperforms the previous PID controller approach \cite{narayanan2021optimizing}.

\begin{figure*}[h]
\vspace{-2em}
    \centering
    \includegraphics[scale=0.1]{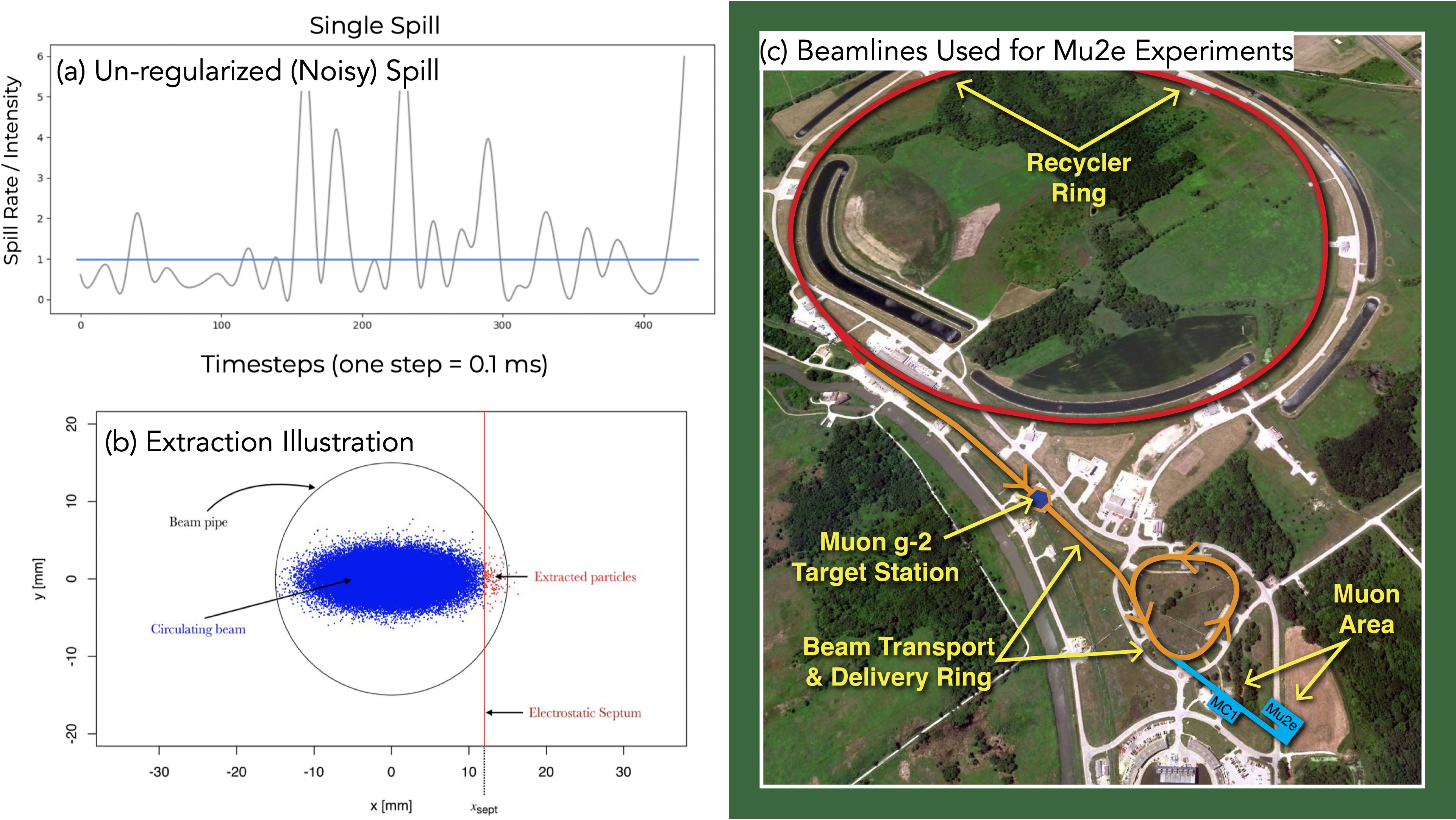}
    \vspace{-0.5em}
    \caption{\small \textbf{(a)}: The extraction (or ‘spill’) of protons from the Delivery Ring is noisy (deviates from 1) without any regulation. \textbf{(b)}: A snapshot of the beam in physical space at the extraction location. As the horizontal beam size increases, a slice of circulating beam (that is past the position of the electrostatic septum) is extracted. \textbf{(c)}: To create the muons, proton pulses are made to hit a production target and muons are obtained from the secondaries. The proton pulses with the required time structure are created by extracting them from an accelerator ring called Delivery Ring at Fermilab and sending it to the Mu2e production target.}
    \label{fig:mu2e}
\end{figure*}

The goal of Mu2e experiment \cite{bartoszek2015mu2e,bernstein2019mu2e,narayanan:ipac21-thpab243,narayanan:napac22-mopa75} at Fermilab is to search for new physics by studying the decay of muons into electrons. 
This intricate experiment places stringent demands on the quality of the proton beam directed at the muon production target, see \cref{fig:mu2e}. 
These requirements are essential to minimize background particle physics processes that could obscure the discovery signal. 
One of the key prerequisites for the Mu2e experiment includes achieving a highly uniform extracted beam intensity during each spill of protons with 8 GeV kinetic energy. 
In order to achieve this, the \textbf{S}pill \textbf{R}egulation \textbf{S}ystem (\textbf{SRS}) \cite{thieme-icfaml22,narayanan2021optimizing,ibrahim2019preliminary} is being developed to govern the extraction process of the beam and mitigate various sources of fluctuations in the spill profile. 

The SRS adjusts the power supply currents of three dedicated fast quadrupoles (resulting in varying its magnetic fields) and this variation results in controlling the variations in the spill intensity (\cref{fig:mu2e}~(c)).
Ideally, the SRS aims for the spill intensity to be perfectly uniform.

Our approach involves utilizing a Proximal Policy Optimization model with a differentiable Mu2e simulator \cite{narayanan2021optimizing} to regulate random generated spills. 
In each episode (or spill), the simulator generates a series of random spills, each with varying intensities (shown in \cref{fig:mu2e}~(a)). 
Subsequently, the PPO model intervenes in each individual time step to correct these generated spills by adjusting the control signal of the Mu2e simulator. 
The primary goal of our model is to bring spill rate of all spills as close to a value of 1 as possible.
To optimize this objective function, while mitigating the influence of excessively high or low intensity spills, we implement an exponential moving average (EMA) to measure the deviation of the sequence (spill rate) from the desired value of 1. 
Additionally, we incorporate neuralized PID controller, encompassing proportional, integral, and derivative components in the state representation. 
By employing different random seeds, our simulator can generate a diverse profile of noisy spills, reflecting the real-world scenarios.

We demonstrate superior performance by numerically benchmarking our methods with PID controller. 
Specifically, our experiments compare SDF (defined in \eqref{eqn:SDF}) performance on different seeds.
Our results show that our methods consistently improve the SDF by 13.6\% for 9 random generated spills and achieve 1.6\% improvements compared to the PID controller.

The rest of this paper commences by a detailed description of our proposed method. Subsequently, we present numerical results to showcase the performance of our approach. Finally, we conclude this paper with a discussion of potential future directions and avenues for further research.

\section{Methodology}
\label{sec:method}
\vspace{-0.05in}
In this section, we first introduce the machine learning's role in optimizing Fermilab's accelerator parameters. Then, we provide a detailed design of our RL-enhanced regularization system.

\subsection{Problem Setup}
\label{ssec:set-up}
\vspace{-0.05in}
Large fluctuations in the proton spill rate result in large fluctuations in the intensity of produced muons. 
And this in turn causes background effects that hinder the signals of new physics.
To combat these, we approach the beam intensity regulation challenge, specifically controlling the spill regulation system, as a tracking control problem. 
The objective is to keep certain signals (spill intensity) close to specific reference values ($\simeq 1$) by controlling the quadrupole currents (power supply of 3 dedicated fast-ramping quadrupoles) in the regulation system. 
By doing so, we adjust the magnetic field, which subsequently adjusts the beam intensity throughout the delivery ring, as depicted in \cref{fig:mu2e}~(c).

\textbf{Objective.}
To handle the constraint challenges posed by Mu2e experiment, we regulate the uniformity of the extracted spill by increasing its Spill Duty Factor (SDF), 
\bea
\label{eqn:SDF}
\text{SDF}\coloneqq1/\left({1+\sigma_{\text{spill}}^2}\right),
\eea
by regulating the extraction process (where $\sigma_{\text{spill}}$ is the standard deviation in the spill rate). 
The ultimate goal for SRS in the Mu2e experiment is to achieve a SDF of 0.6 or higher, with an ideal spill having a constant spill rate value of 1 and an SDF of 1.

\textbf{Mu2e Simulator.} We employ a differentiable simulator proposed in \cite{narayanan2021optimizing} to replicate the beam physics process in Mu2e experiments realistically. 
This physics simulator generates spill intensity and the associated data. 
It subsequently conveys this data to the RL agent, which aids in training the RL model. 
Once trained, the RL model transmits control signals back to the simulator, allowing it to regulate the \textit{deviation} in the spill rate based on these signals and provide the modified data to the RL agent.
This process is shown in \cref{fig:method}~(a).

\subsection{Proximal Policy Optimization (PPO) Controller for Spill Regulation System}
\vspace{-0.05in}
\begin{figure*}[t]
\vspace{-2em}
    \centering
    \includegraphics[width=\textwidth]{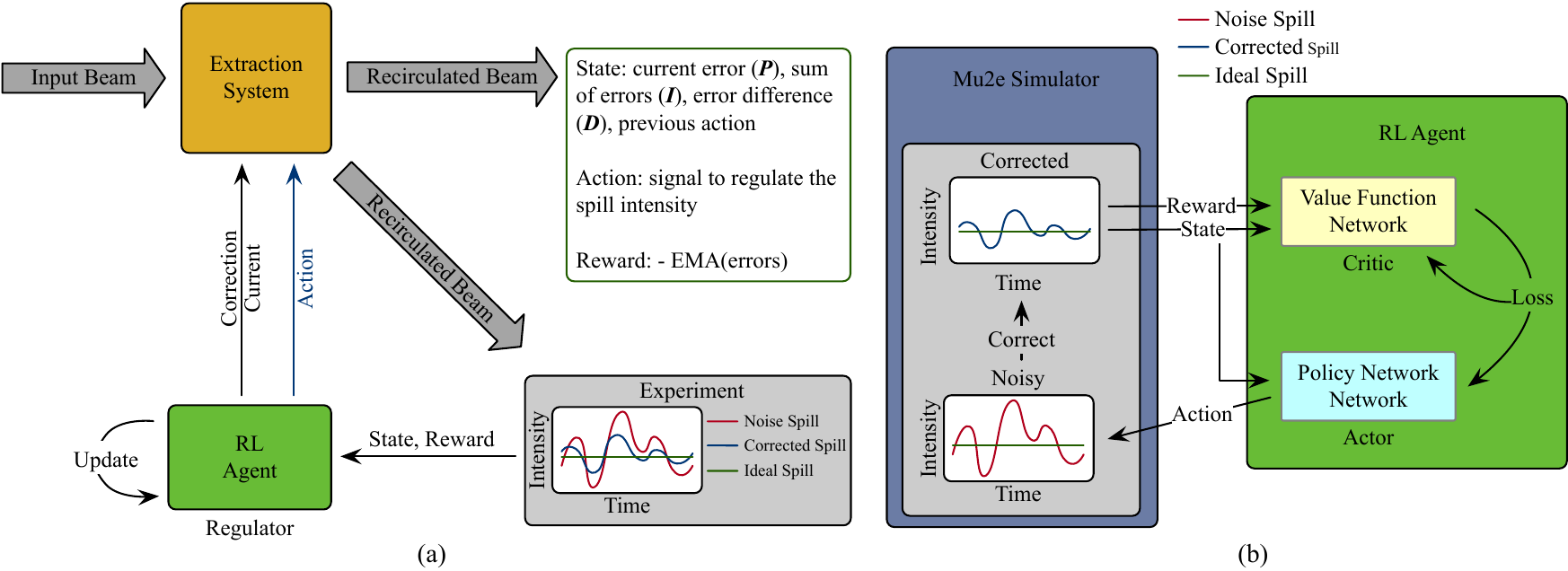}
    \vspace{-1.5em}
    \caption{\small 
    \textbf{(a)}:  The Mu2e simulator initially generates the noised spill data. 
    The agent proceeds to adjust the spill and the code employs this adjusted spill to compute relevant information such as the state and reward. 
    These pieces of information are instrumental in training the RL agent, which in turn offers new actions for the subsequent time step to refine the spill.
    \textbf{(b)}: 
    The simulator refines (corrects) the spill derived from noisy data. 
    It conveys the state and reward, calculated using the corrected spill, to update the value network responsible for evaluating the quality of the correction. 
    Subsequently, the state and loss generated by the value network contribute to the adaptation of the policy network. 
    The policy network, in response, generates new actions for spill regulation.
    }
    \label{fig:method}
\end{figure*}

\textbf{Reward Function.}
Let $x_t$ be the observation of one single spill signal at time step $t$ and $\sigma = 1$ be the corresponding target reference value.
The reward function at $t$ is defined as the exponential moving average
\bea
r_t=-{\rm EMA}(t,\alpha), \alpha \in \[0, 1\],\;\text{where}\;\;{\rm EMA}(t,\alpha) = \alpha|x_t - \sigma| + (1-\alpha){\rm EMA}(t-1,\alpha).
\eea
EMA gives more weight to recent spills and less weight to older spills. This helps in reducing the impact of short-term fluctuations in the spills, making it easier to identify underlying trends.

\textbf{PID Controller.}
A PID controller in discrete-time operation captures past details regarding tracking errors, their integrals, and derivatives within a linear control strategy.
We denote the time series of spill signal at time $t$ as $o_t = (x_0, x_1, \cdots, x_t)$.
In formal terms, the discrete-time PID controller's policy, characterized by its parameters $K_P$, $K_I$, and $K_D$, is expressed as:
\vspace{-0.1truein}
\bea
\label{eqn:pid}
\pi^{\text{PID}}(o_t)
=K_P\(x_t-\sigma\)
+K_I \sum_{\tau=0}^t\(x_{\tau}-\sigma\)
+K_D\frac{\(x_t-\sigma\)-\(x_{t-1}-\sigma\)}{\Delta  t },
\vspace{-0.2truein}
\eea
where $K_P,K_I,K_D$ are tuneable scalar coefficients and $\Delta t$ is the discrete time interval.

\textbf{Model: PPO with Neuralized PID Policy.}
Our model leverages the inductive bias of PID (Proportional-Integral-Derivative) controller, and propose a neuralized PID policy function. 
Specifically, we incorporate tracking errors, integrals, and derivatives as components of the state vector $s_t$. Furthermore, we employ a linear network to parametrize the standard PID controller \eqref{eqn:pid}, and use it as a part of our policy function.
Therefore,
the policy function of our model not only enables the extraction of external information (based on previous actions) but also includes the PID control signals $K_P$, $K_I$, and $K_D$ as its learnable parameters. 
Remarkably, when learned effectively, our policy network outperforms the standard PID controller \eqref{eqn:pid}, making it a highly adaptable solution.

Our approach incorporates three key components: the PPO algorithm, the EMA reward function, and the neuralized PID controller. 
The PPO algorithm \cite{schulman2017proximal}, a reinforcement learning technique, plays a central role in optimizing policy functions to enhance decision-making in sequential tasks and we specifically chose PPO to refine our reward function. 
Rather than relying on a single spill for reward computation, we employ EMA within the reward function. 
This approach allows us to both capture trends across a sequence of spills and mitigate fluctuations.
Additionally, we integrate the neuralized PID bias into our policy network. 
In this setup, our policy network consists of a trainable PID controller and a linear projection of past actions.
Let's denote the state at time step $t$ as $s_t$, the action as $a_t$.
We combine the PID policy $\pi^{\text{PID}}$ and action policy $\pi^{\text{action}}$ to formulate RL policies as $\pi$.
The variable $a_t$ represents the control signal used for regulating the spill.
The state $s_t$ encompasses both the previous action $a_{t-1}$ and the time series of the spill signal $o_t$.
As a result, the action at time $t$, $a_t$, can be represented as follows:
\bea
a_t = \pi(s_t) = \pi^{\text{PID}}(o_t) + \pi^{\text{action}}(a_{t-1}).
\eea
The learning process is shown in \cref{fig:method}~(b).

\section{Experimental Studies}
\label{sec:exp}
\vspace{-0.05in}
We validate our method by using the Mu2e differentialble simulator \cite{narayanan2021optimizing} as the RL environment.
Our experiments involve the utilization of both the Mu2e simulator and our RL-enhanced spill regularization system.

\begin{figure*}[t]
\vspace{-2em}
    \centering
    \includegraphics[width=\textwidth]{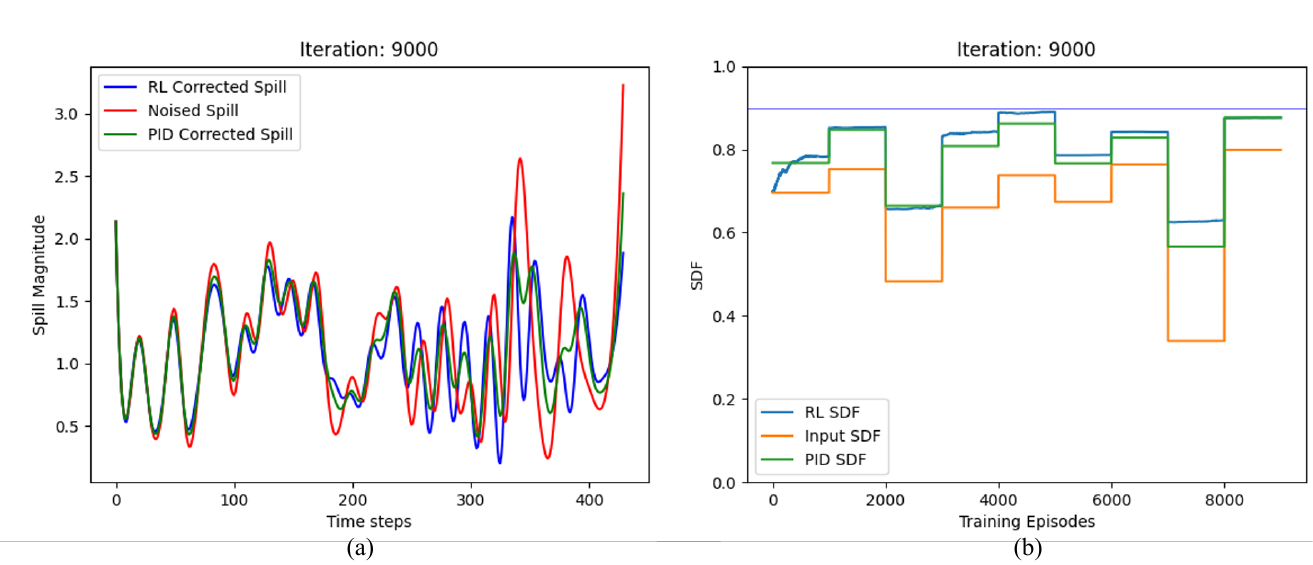}
    \vspace{-1.5em}
    \caption{\small 
    Spill intensity and SDF comparison in different seeds. 
    \textbf{(LHS)}: Comparison of Spill Intensity: The spill intensity corrected by RL is closer to 1 when compared to the PID-corrected spill.
    \textbf{(RHS)}: Comparison of SDF: After 600 training iterations, the SDF achieved by RL outperforms or nears the SDF obtained through PID.
    }
    \label{fig:results}
\end{figure*}

\textbf{Settings.}
We configure various hyperparameters for both the simulator and the RL agent in our experiment. 
Specifically, we configure the simulator to generate 430 spills per iteration, equivalent to 10 data points per millisecond within a 43 ms spill duration, aligning closely with realistic settings as detailed in \cite{narayanan2021optimizing}. 
In terms of reward calculation, we choose a value of $\alpha = 0.5$ for the EMA component.
Regarding the RL agent, we employ the \texttt{\textbf{stable-baselines3}}\footnote{\url{https://github.com/DLR-RM/stable-baselines3}} PPO model \cite{raffin2021stable}. The actor network is designed as a single linear layer, while the critic network took the form of a two-hidden layer ($64 \times 64$) MLP network. 
The learning rate is set at $1 \cross 10^{-4}$. 
We change the random seed every 1000 (spills) epochs.

\textbf{Results.}
In \cref{fig:results}, we examine the spill intensity in scenarios involving unregularized, PID-regularized, and RL-regularized setups. \cref{fig:results}~(a) demonstrates that the spill corrected by RL brings the spill rate closer to the ideal value of 1 when compared to the PID-corrected spill, on the first random seed configuration.
Furthermore, we provide a visual representation of the evolution of the SDF during the training process. \cref{fig:results}~(b) illustrates that the SDF achieved through RL uniformly surpasses (or approaches) that of PID regulation after 600 episodes.

\textbf{Ablation Study.} 
We systematically vary parameters and model architectural configurations to differentiate their impact in these four aspects:
1), we compare the usage of PID controller and a neural network in Policy network's;
2), we experiment with various values of the smoothing factor ($\alpha$) in the Exponential Moving Average (EMA); 
3), we explore the effect of employing different reward functions;
4), we select alternative states from the Mu2e (spills) environment to replace the usage of inductive biases from the PID;
5), we compare two RL algorthims: Soft-Actor Critic (SAC) \cite{DBLP:journals/corr/abs-1801-01290} and PPO.
For all the experiments, we use the same random seed series. 
We report the average improvements of SDF of our proposed methods, compared to PID regularized SDF and unregularized SDF in  \cref{tab:ablation}.

\begin{table}[]
    \caption{Evaluation of the impact of various parameters. ``v.s. PID'' represents the SDF improvement by comparing our methods to PID method. ``v.s. Noise'' is the SDF improvements of our methods compared to unregularized SDF. 
    NN: neural network policy; PID: neuralized PID policy; EMA: exponential moving average of errors; $\alpha$: EMA's smooth parameter; SUM: sum of errors; PPO: Proximal Policy Optimization; SAC: Soft-Actor Critic; P: current error; I: sum of errors; D: error difference; Act: previous action; CD: corrected spill difference; Over-1: number of noisy spill intensity $\geq$ 1. }
    \vspace{0.1em}
    \centering
    \resizebox{0.8\textwidth}{!}{%

    \begin{tabular}{cccccccc}
    \toprule
    v.s. PID  & v.s. Noise & Policy Network & Reward Func. & RL Algorthim & State \\
    \midrule
    -4.11 & 7.90  & PID & -EMA ($\alpha$=0.1) & PPO & P, I, D, Act \\
    -11.12 & 0.89  & NN  & -EMA ($\alpha$=0.5) & PPO & P, I, D, Act \\
     1.42 & 13.55 & PID & -EMA ($\alpha$=0.9) & PPO & P, I, D, Act \\
    1.95 & -10.06  & PID & -SUM                & PPO & P, I, D, Act \\
    1.46 & 13.47  & PID & -EMA ($\alpha$=0.5) & PPO & P, I, D \\
    5.76 & -6.25  & PID & -EMA ($\alpha$=0.5) & PPO & CD, Over-1, P, Act \\
    -12.60 & -24.61 & PID & -EMA ($\alpha$=0.5) & SAC & P, I, D, Act \\
    \midrule
    \textbf{1.65} & \textbf{13.67}  & PID & -EMA ($\alpha$=0.5) & PPO & P, I, D, Act \\
    \bottomrule
    \end{tabular}
    }
    
    \label{tab:ablation}
\end{table}

\section{Related Works in Real-time Edge AI for Distributed Systems (READS)}
\label{sec:related}
\vspace{-0.05in}
The READS project~\cite{seiya2021accelerator} includes two primary sub-projects: 1) Beam Loss Deblending for the Main Injector and Recycler, and 2) Mu2e Spill Regulation, which is the subject of this work.

\textbf{Beam Loss Deblending for Main Injector and Recycler.} 
In this sub-project, the challenge is distinguishing between beam losses from two adjacent accelerators (Recycler Ring and Main Injector) that share the same monitoring system. 
We summarize some related works in resolving this challenge bellow. 
The Deblending model (DBLN) \cite{hazelwood2021real} uses a dual-network MLP setup to classify beam losses and predict their probabilities for each accelerator and beam loss monitor (BLM).
The Many Models \cite{hazelwood-icalepcs23} employs individual MLP models for each BLM and aggregates their outputs to enhance local pattern recognition for beam losses. 
The semantic regression model \cite{thieme2022semantic} uses the U-Net architecture to capture each accelerator's localized and extensive beam loss patterns. 
As for hardware, a series of work on Field Programmable Gate Array (FPGA)-based edge-AI systems \cite{berlioz:napac22-mopa15, ibrahim2023fpga, shi2023ml, arnold2023edge} are developed and deployed for real-time identification of beam loss sources in accelerator complex, enhancing operational accuracy and efficiency.

\textbf{Mu2e Spill Regulation.} There are some related works in solving challenges (as described in Section \ref{ssec:set-up}) in Mu2e Spill Regulation project. 
The PID method \cite{narayanan2021optimizing} employs a hybrid machine learning approach that integrates a neural network (NN) to optimize PID controller gains within an end-to-end machine learning (ML) differentiable simulator simulator, thereby enhancing the spill quality of a slow extraction system for proton delivery.
The GRU network \cite{narayanan:napac22-mopa75} ingests a history of spill intensity observations as input and makes predictions for the quadrupole adjustments based on these inputs.

\section{Conclusion}
\label{sec:conclusion}
\vspace{-0.05in}
We present an innovative RL-enhanced spill control system, utilizing a neural PID controller as the policy function, to tackle beam regulation issues in Mu2e experiments. 
To simulate real-world spill control scenarios, we utilize a differentiable Mu2e simulator to create spills, improve spill adjustments, and establish reward signals. 
Furthermore, we harness an RL-based controller for fine-tuning control signals during the regularization process. 
Our approach outperforms the PID-based regularization model, achieving a 1.6\% higher SDF performance, with a 13.6\% improvement in the SDF of unregularized spills, confirmed across nine different settings with random seeds.

\appendix	
\label{sec:append}
\section{Future Directions}
We outline two potential directions for future investigations toward a \textit{model-free} RL controller for the Mu2e experiments.

\paragraph{Pretrained Transformers.}
A main motivation for our method is the sequential modeling of noisy spill rate hard to converge for naive RL algorithms.
This challenge persists even with the simplified simulator environment \cite{narayanan2021optimizing}.
While our approach incorporates model-based inductive bias (such as PID), another popular method in the literature involves using transformers \cite{vaswani2017attention} for generative trajectory modeling in RL \cite{paischer2022history,janner2021offline,chen2021decision}.
This approach is beneficial for our problem: 
it scales up easily using existing transformer-based language and vision models like BERT \cite{zhou2023dnabert,ji2021dnabert,devlin2018bert} and GPT \cite{wei2021finetuned}, enables stable training \cite{zhai2023stabilizing}, and addresses the challenges of short-sighted behavior and long-term credit assignment in RL \cite{chen2021decision}.

However, a major issue with training transformers solely on observed samples from the environment is their tendency to overfit, especially given the typically limited data generated by the current policy \cite{chen2021decision}. 
To circumvent this, we propose using a Transformer pre-trained on extensive observations/samples without any fine-tuning or weight adjustments \cite{paischer2022history}. 
Modern Hopfield networks \cite{hu2023sparse,paischer2022history,ramsauer2020hopfield} provide a natural approach for utilizing frozen transformers via their associative memory interpretation.
We notice that there are works  in (i) employing ``frozen'' modern Hopfield networks with pre-trained transformers for RL \cite{paischer2022history}, and (ii) addressing noisy sequence modeling using sparse modern Hopfield models \cite{hu2023sparse}. 
We  would like to explore the integration of (i) \& (ii)  in the future.

\paragraph{Feature Programming.}

A key innovation in our approach is the integration of \textit{model-based} inductive bias from PID controllers into our policy network's architecture.
This involves manually crafting features (PID parameters) from sequences of spills. 
However, this method risks missing other crucial features and may introduce instability in the training process, as evidenced in \cref{tab:ablation}. 
To more effectively identify and leverage informative features for RL model training, we propose employing Feature Programming \cite{reneau2023feature}, an automated feature engineering framework for noisy multivariate time series. 
This framework will be used to (i) generate informative features from noisy spill sequences and (ii) incorporate model-based inductive bias from PID controllers. 
In future work, we aim to combine both (i) and (ii) to develop a \textit{model-free} RL controller for the Mu2e experiments.

\clearpage
\section*{Acknowledgments}
JH and CW would like to thank Jiayi Wang for facilitating experimental deployments.
The authors would like to thank the anonymous reviewers and program chairs for constructive comments.

AN was previously affiliated with Northern Illinois University for the earlier part of the READS collaboration, DeKalb, IL, USA.
Besides the aforementioned support, 
JH is partially supported by the Walter P. Murphy Fellowship.
HL is partially supported by NIH R01LM1372201, NSF CAREER1841569, DOE DE-AC02-07CH11359 and a NSF TRIPODS1740735.
This research was supported in part through the computational resources and staff contributions provided for the Quest high performance computing facility at Northwestern University which is jointly supported by the Office of the Provost, the Office for Research, and Northwestern University Information Technology.
The content is solely the responsibility of the authors and does not necessarily represent the official views of the funding agencies.

\bibliographystyle{plainnat}
\bibliography{refs}

\end{document}